\documentclass{article}

\usepackage{arxiv}

\usepackage[utf8]{inputenc} 
\usepackage[T1]{fontenc}    
\usepackage{hyperref}       
\usepackage{url}            
\usepackage{booktabs}       
\usepackage{amsfonts}       
\usepackage{nicefrac}       
\usepackage{microtype}      
\usepackage{cleveref}       
\usepackage{lipsum}         
\usepackage{graphicx}
\usepackage{natbib}
\usepackage{doi}
\usepackage{graphicx}
\usepackage{xcolor}
\usepackage[format=plain,
            labelfont={bf},
            textfont=it]{caption}
\usepackage{array}
\usepackage[english]{babel}

\title{SUTRA: Scalable Multilingual Language Model Architecture}

\newif\ifuniqueAffiliation
\uniqueAffiliationtrue

\ifuniqueAffiliation 
\author{ {\hspace{1mm}Abhijit Bendale}\thanks{Correspondence can be addressed to \texttt{abhijit@two.ai}} \\
	Two Platforms\\
	\texttt{abhijit@two.ai} \\
	\And
	\hspace{1mm}Michael Sapienza \\
	Two Platforms\\
	\texttt{michael@two.ai} \\
	\And
	\hspace{1mm}Steven Ripplinger \\
	Two Platforms\\
	\texttt{steven@two.ai} \\
	\And
	\hspace{1mm}Simon Gibbs \\
	Two Platforms\\
	\texttt{simon@two.ai} \\
	\And
	\hspace{1mm}Jaewon Lee \\
	Two Platforms\\
	\texttt{jaewon@two.ai} \\
	\And
	\hspace{1mm}Pranav Mistry \\
	Two Platforms\\
	\texttt{pranav@two.ai} \\
}
\else
\usepackage{authblk}

\author[1]{%
	{\usebox{\orcid}\hspace{1mm}Abhijit Bendale\thanks{\texttt{abhijit@two.ai}}}%
}
\author[1,2]{%
	{\usebox{\orcid}\hspace{1mm}Pranav Mistry\thanks{\texttt{pranav@two.ai}}}%
}
\affil[1]{TWO AI}
\affil[2]{TWO AI}
\fi

\renewcommand{\shorttitle}{TWO Technical Report}

\hypersetup{
pdftitle={A template for the arxiv style},
pdfsubject={q-bio.NC, q-bio.QM},
pdfauthor={Abhijit Bendale, Pranav Mistry},
pdfkeywords={First keyword, Second keyword, More},
}

\begin{document}

\maketitle

\begin{abstract}

In this paper, we introduce SUTRA, multilingual Large Language Model architecture capable of understanding, reasoning, and generating text in over 50 languages.
SUTRA's design uniquely decouples core conceptual understanding from language-specific processing, which facilitates scalable and efficient multilingual alignment and learning.
Employing a Mixture of Experts framework both in language and concept processing, SUTRA demonstrates both computational efficiency and responsiveness.
Through extensive evaluations, SUTRA is demonstrated to surpass existing models like GPT-3.5, Llama2 by 20-30\% on leading Massive Multitask Language Understanding (MMLU) benchmarks for multilingual tasks.
SUTRA models are also online LLMs that can use knowledge from the internet to provide hallucination-free, factual and up-to-date responses while retaining their multilingual capabilities.
Furthermore, we explore the broader implications of its architecture for the future of multilingual AI, highlighting its potential to democratize access to AI technology globally and to improve the equity and utility of AI in regions with predominantly non-English languages.
Our findings suggest that SUTRA not only fills pivotal gaps in multilingual model capabilities but also establishes a new benchmark for operational efficiency and scalability in AI applications.

\end{abstract}


\section{Introduction}

\begin{figure}[tb]
\centering
\includegraphics[width=1.0\linewidth]{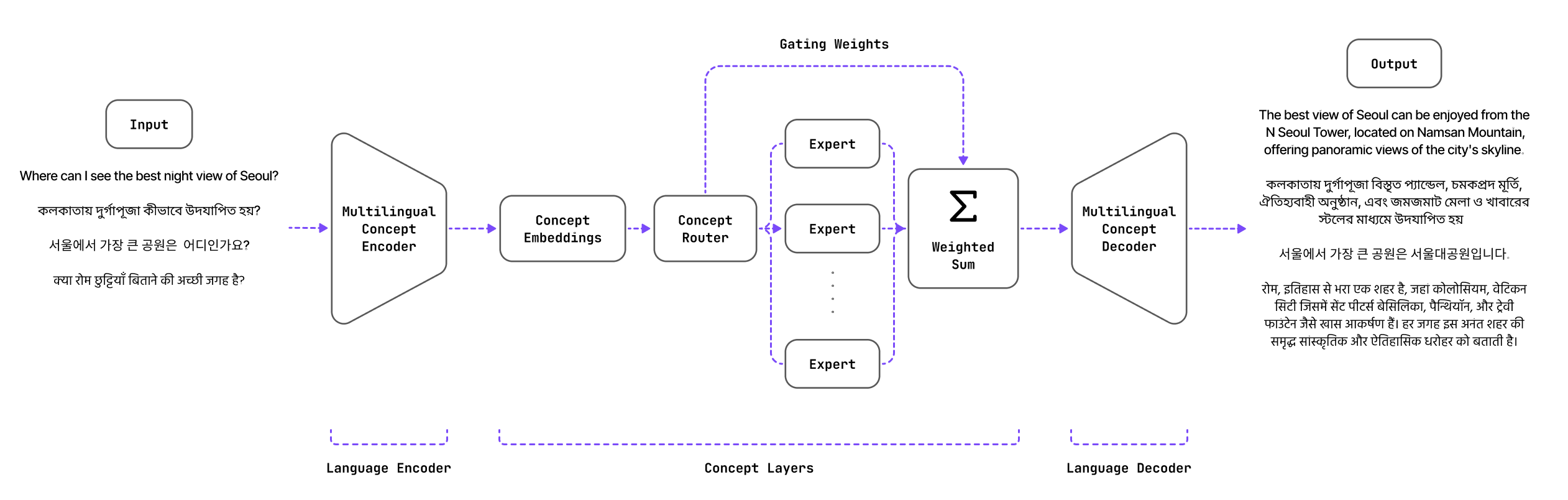}
\caption{SUTRA is a novel multilingual large language model architecture that is trained by decoupling concept learning from language learning.
  The input is processed through a multilingual concept encoder, followed by the concept model and finally through a multilingual concept decoder to generate the output response.}
\label{fig:inference_pass}
\end{figure}

Recent advancements in Large Language Models (LLMs) have predominantly focused on a limited set of data-rich languages, with training datasets being notably skewed towards English.
This skew results in a significant bias, rendering LLMs less capable of understanding, processing, and generating text in languages with substantial speaker populations.
Notably, languages such as Hindi, Arabic, Bengali, and Japanese, each with over 250 million speakers, constitute less than 3\% of the data typically used for training these models.
This enduring bias represents a substantial challenge, as it is not feasible for LLMs to underperform in languages spoken by vast numbers of people.
In light of these disparities, our work seeks to bridge the gap between market demands and the current capabilities of LLMs.
We strive to mitigate linguistic inequality inherent in recent multilingual Instruction-Following Task (IFT) models.
Our goal is to develop a model adept at performing downstream tasks in any supported language, eliminating the need for multilingual speakers to default to English for prompts. 

The advent of Large Language Models (LLMs) like GPT-3.5, BERT, and others has revolutionized the field of artificial intelligence, offering unprecedented capabilities in natural language understanding and generation \citep{brown2020language, devlin2018bert}.
These models have been instrumental in a variety of applications, ranging from conversational agents to complex decision support systems.
However, the vast majority of these models predominantly cater to English, which is not only limiting in terms of linguistic diversity but also in accessibility and utility across different geographic and cultural contexts \citep{jia2019bias}.

Addressing the challenge, multilingual LLMs have been developed, but these models often suffer from significant trade-offs between performance, efficiency, and scalability, particularly when extending support across a broader spectrum of languages \citep{conneau2020unsupervised}.
The most common approach has been to train large universal models capable of understanding multiple languages.
Yet, these models, such as BLOOM and Llama2, typically underperform in languages that are less represented in the training data due to the difficulty of balancing language-specific nuances \citep{smith2021can, zhang2020improving}.
The development of SUTRA was motivated by the inherent limitations in existing multilingual LLMs.
On the one hand there are language-specific LLMs like HyperClova in Korean or OpenHaathi in Hindi.
Scaling and managing such models is not only costly, but challenging due to the exponential data and training requirements.
Each time a new base model is created, it would require fine-tuning for many different languages.
On the other hand large traditional LLMs like BLOOM and Llama2 struggle on multilingual tasks, as they have to balance learning core multilingual capabilities and skills, often resulting in confusion between languages.
For example, when asking GPT a question in Korean, one might notice how formal and informal tones are often misplaced.
SUTRA was developed to address two main challenges of existing multilingual LLMs: the high computational/scaling costs of language-specific models, and the difficulties larger models face with multilingual tasks (leading to language confusion).

In response to these limitations, we introduce SUTRA (Sanskrit for "thread"), a transformative approach in the architecture of multilingual LLMs.
SUTRA uniquely separates the process of concept learning from language learning, as illustrated in Figure~\ref{fig:inference_pass}.
SUTRA is a novel multilingual large language model architecture that is trained by decoupling concept learning from language learning.
This architecture enables the core model to focus on universal language-agnostic concepts while leveraging specialized neural machine translation (NMT) mechanisms for language-specific processing, thus preserving linguistic nuances without compromising the model's scalability or performance \citep{wu2019google}.
SUTRA employs a Mixture of Experts (MoE) strategy, enhancing the model's efficiency by engaging only the relevant experts based on the linguistic task at hand \citep{shazeer2017outrageously}.
Furthermore, SUTRA models are internet-connected and hallucination-free models that understand queries, browse the web, and summarize information to provide the most current answers, without loosing their multilingual capabilities.
A combination of multilingual skills, online connectivity, and efficiency in language generation incorporated by SUTRA models promises to redefine the landscape of multilingual language modeling.

In the subsequent sections, we will outline the architecture of SUTRA, our training methodology, and present a comprehensive evaluation that demonstrates its superiority over contemporary multilingual models on several benchmarks, including the Massive Multitask Language Understanding (MMLU) tasks \citep{hendrycks2021measuring}.
By effectively decoupling concept learning from language processing, SUTRA sets a new paradigm in the development of LLMs, promising broader accessibility and enhanced performance across diverse linguistic landscapes.

The paper is organized as follows: First, we discuss related work in the context of SUTRA. Next, we describe the architecture and training methodology adopted. We then discuss the data used for training and provide both an evaluation of SUTRA's multilingual as well as online cabailities.
Finally, we discuss how to build more inclusive LLMs for the benefit of a wider community.

\section{Related Work}

\textbf{Large Language Models \& Multilinguality:} The field of Large Language Models (LLMs) has witnessed substantial advancements, particularly through the development of models such as GPT-3 \citep{brown2020language} and BERT \citep{devlin2018bert}, which have set new benchmarks in language understanding and generation. These models utilize vast amounts of data to learn complex patterns and generate coherent text, but their primary limitation has been a focus largely on English language data.
In response to the need for supporting global linguistic diversity, research has expanded into multilingual LLMs.
Pioneering works like mBERT \citep{devlin2018bert} and XLM-R \citep{conneau2020unsupervised} have demonstrated significant potential in learning representations that generalize across languages.
However, these models often face challenges in balancing performance across languages, especially for those less represented in training datasets \citep{conneau2020unsupervised}. Further, as the number of languages increases, the scalability and efficiency of these models often degrade, necessitating more specialized architectures to handle the diversity of languages effectively \citep{smith2021can}.

\textbf{Neural Machine Translation:} Neural Machine Translation (NMT) has been integral to the progress in multilingual model performance. Early NMT systems were limited by the complexity of their architectures and the quality of their translations, especially in low-resource languages \citep{wu2019google}. Recent studies have revisited the core challenges of machine translation in the context of advanced Large Language Models (LLMs). The work by \citet{koehn2017six} offers insights into the ongoing relevance of challenges such as domain mismatch, rare word prediction, and translation of long sentences, even as LLMs have shown significant improvements in these areas. Additionally, a study by \citet{son2023translation} explored the translation performance of LLMs from the user's perspective, highlighting their potential to enhance the translation of long sentences while also identifying persistent challenges around domain mismatch and rare word prediction. The work by \citet{wu2016google} on Google's neural machine translation system has also served as a benchmark for progress in this field, bridging the gap between human and machine translation. Recently, the work by \citet{costa2022no} showed that the Mixture of Experts architecture can be used effectively in the context of Neural Machine Translation and have considerable gains in translation performance on various low-resource languages.

\textbf{Mixture of Experts:} Mixture of Experts (MoE) has emerged as a promising architecture for managing the computational costs associated with scaling up large language models (LLMs). Recent studies have explored the benefits of MoE in this context. \citet{zhou2022mixture} proposed a Mixture-of-Experts with Expert Choice Routing, which enables dynamic allocation of data among different experts, allowing each expert to focus on its expertise and achieve model sparsity. Similarly, \citet{zoph2022designing} investigated the design of effective sparse expert models, highlighting the importance of carefully balancing the number and size of experts to optimize performance. Additionally, \citet{ott2022opt} introduced the OPT family of open pre-trained transformer language models, which leverage MoE to achieve significant improvements in efficiency and scalability compared to dense models. Furthermore, \citet{zheng2019chid} explored the application of MoE in the context of Chinese idiom datasets, demonstrating the potential of this approach to enhance language understanding tasks. These studies collectively suggest that MoE can serve as an effective choice for building highly capable and computationally efficient LLMs.

\textbf{Multimodal LLMs:} Researchers have also explored the potential of multimodal Large Language Models that can process and generate content across different modalities, such as text, images, and video. For example, the work by \citet{dai2019transformer} has investigated the use of multimodal models for tasks like image captioning and visual question answering, demonstrating their ability to leverage cross-modal information to enhance performance. Similarly, the study by \citet{nichols2008tutorial} has explored the application of multimodal models in the context of computational linguistic phylogeny, highlighting their potential to uncover insights from diverse data sources. Additionally, the recent advancements in the field of multimodal machine translation, as discussed by \citet{birch2021neural}, have shown the benefits of integrating visual information into language models to improve translation quality.

\begin{table}[ht]
\setlength{\abovecaptionskip}{10pt} 
\setlength{\belowcaptionskip}{5pt} 
\centering
\begin{tabular}{lllll}
\toprule
\textbf{Model Name}   & \textbf{Creator}     & \textbf{License}   & \textbf{Context Window} & \textbf{Training Cut-off Date} \\ \midrule
GPT-4                 & OpenAI               & Proprietary        & 8k                      & Sep 2021                      \\
GPT-3.5 Turbo         & OpenAI               & Proprietary        & 16k                     & Sep 2021                      \\
Llama 2 Chat (70B)    & Meta                 & Open               & 4k                      & Sep 2022                      \\
Mixtral 8x7B Instruct & Mistral              & Open               & 32K                     & Sep 2023                      \\
SUTRA                 & Two Platforms        & Proprietary        & 32K                     & Up-to-Date                    \\ \bottomrule
\end{tabular}
\caption{Comparison of various AI models for their knowledge cut-off dates. The knowledge cutoff represents the latest point at which the language model was updated with new information, beyond which it lacks any further data or recent developments. Online models like SUTRA have the ability to continuously learn and reason from recent data.}
\label{table:ai_models}
\end{table}

\textbf{Online LLMs:} Modern Large Language Models like Llama2, GPT-3.5, and GPT-4 have been engineered as comprehensive, open-domain chatbots capable of engaging in extended dialogues on a variety of topics. Yet, they face a significant limitation: their data is time-locked, leading to a cutoff date for knowledge. Due to this, these models sometimes generate responses that are plausible yet factually incorrect, diminishing the reliability of their output as noted by \citet{vu2023freshllms} and \citet{press2022measuring} and such inaccuracies are often linked to outdated information embedded in the model's parameters. A detailed list of knowledge cutoff dates for major models is shown in Table \ref{table:ai_models}. While this can be somewhat rectified through additional training with human feedback or by incorporating knowledge-intensive tasks, scaling these solutions to accommodate real-time updates, such as changes in stock prices, remains challenging \citep{komeili2021internet}. In-context learning presents a promising alternative, allowing for the incorporation of real-time data directly into the model's prompts to guide response generation. Although there are ongoing efforts to enhance LLMs with internet search results, effectively leveraging this external data to improve the accuracy of LLM outputs is still under development. In this context, SUTRA stands out by presenting a structured approach for response augmentation, providing the ability to learn, reason, and interpret information from various knowledge sources.

\section{SUTRA Approach}

\subsection{What is SUTRA?}
SUTRA is a novel multilingual large language model architecture that is trained by decoupling concept learning from language learning.
Inspired by how humans learn, SUTRA decouples core concept learning from language learning, making it scalable and easier to reach large number of languages.
Humans first understand the world through concepts and then gradually learn their native language.
Once fluent in one language, they learn new languages without having to re-learn common core concepts.
Similarly, central to our approach is the innovative strategy of separating concept learning from language learning.
This enables the core LLM capabilities to operate within a conceptual or latent space, while the heavy lifting of tokenization and translation is handled by specialized encoders and decoders inspired from Neural Machine Translation.
This approach makes training of LLMs more scalable, whilst making it easier to reach a larger number of languages.

Our training methodology unfolds in three phases: concept learning, language learning and language alignment. 
\begin{itemize}
\item \textbf{Concept Learning:} Initially, the core concept model undergoes training to grasp concepts within a small set of languages, setting a solid foundation for understanding basic concepts and skills. 
\item \textbf{Language Learning:} In parallel, we train specialized Neural Machine Translation (NMT) based encoders and decoders, alongside a multilingual tokenizer, specifically designed to master multi-language translation and ensure concept consistency across languages. 
\item \textbf{Language Alignment:} Finally, we perform language alignment, merging concept understanding with linguistic proficiency. 
\end{itemize}
In the inference stage, SUTRA employs a structured path: Input is processed through an NMT Encoder, followed by the Concept Model, and finally through the NMT Decoder to produce the output.

\subsection{Architecture}

\begin{figure*}
\centering
\vspace{-12pt}
\includegraphics[width=1.0\linewidth,keepaspectratio]{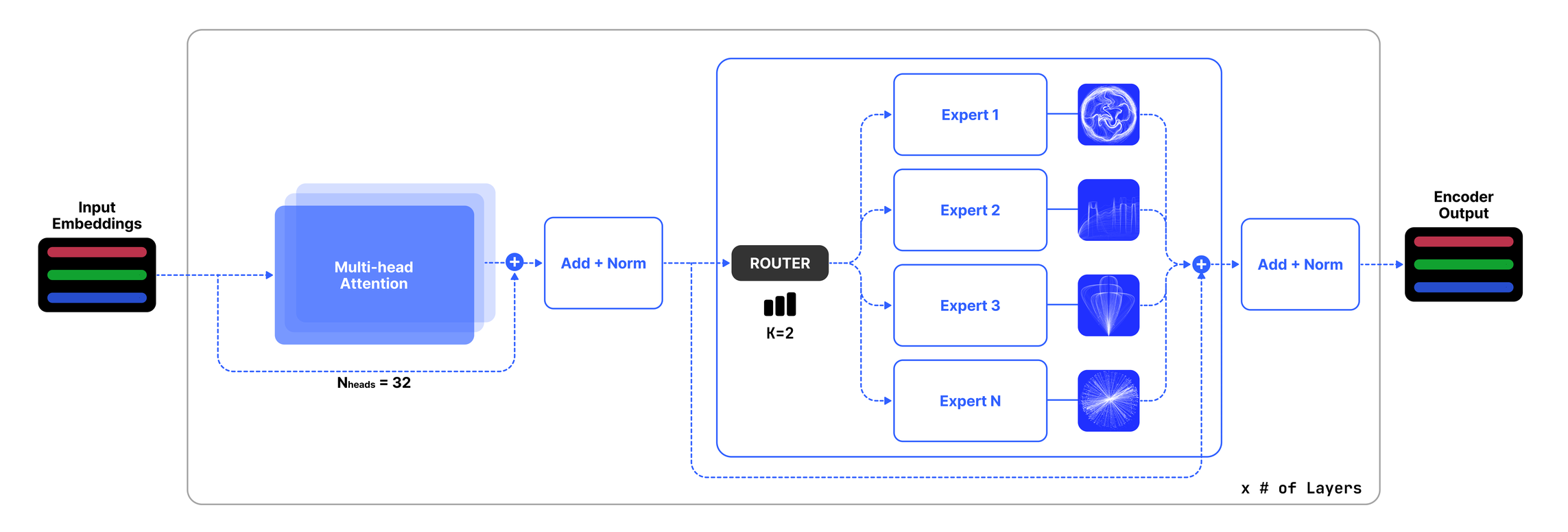}
\caption{\small \textbf{Expert Mixture Layer Configuration.} Input vectors are routed to a subset of the available experts, specifically 2 out of 8, by a specialized router. The aggregate output of this layer is the sum of the individual outputs, each weighted accordingly. Each expert comprises a feedforward module similar to those found in conventional transformer models.}
\label{fig:expert_mixture}
\vspace{-2pt}
\end{figure*}

The architecture of our model, referred to herein as SUTRA, is built upon the foundational principles of the transformer architecture as delineated by \citet{vaswani2017attention}.
Our model retains the enhancements specified by \citet{jiang2023mistral}, with the critical adaptation that it facilitates an extended dense context length of up to 32k tokens.
Moreover, we have employed MoE layers, enabling selective activation of experts and leading to efficiency in computation and memory consumption, as shown in Figure~\ref{fig:expert_mixture}.
The key architectural parameters of SUTRA are encapsulated in Table~\ref{table:model_parameters}.

\begin{table}[tb]
\setlength{\abovecaptionskip}{10pt} 
\setlength{\belowcaptionskip}{5pt} 
\centering
\begin{tabular}{llll}
\toprule
\textbf{Parameter Name} & \textbf{Parameter Value} & \textbf{Parameter Name} & \textbf{Parameter Value} \\ \midrule
Model Dim               & 1024                     & Context Window          & 8K, 32K                  \\
LLM Layers              & 32                       & Batch Size              & 1M Tokens                \\
Attention Heads         & 32                       & FFN Dim                 & 14336                    \\
\# of Experts           & 8                        & Language Enc. Attn Heads & 16                     \\
\# of top Experts       & 2                        & Language Dec. Attn Heads & 16                     \\ \bottomrule
\end{tabular}
\caption{The above table shows some selected model parameters for SUTRA.}
\label{table:model_parameters}
\end{table}

Given an input $x$, the output yielded by the Expert Mixture module is the sum of each expert network's contribution, modulated by the gating network.
Formally, for $ n $ experts $\{E_0, E_1, ..., E_{n-1}\}$, the resultant output is:
\vspace{-5pt}
\begin{equation}
\centering
\sum_{i=0}^{n-1} G(x)_i \cdot E_i(x)
\end{equation}
where $ G(x)_i $ represents the gating function's output, producing an $ n $-dimensional vector corresponding to the $ i $-th expert's activation, while $ E_i(x) $ delineates the $ i $-th expert network's output.
The model capitalizes on sparsity by disregarding inactive experts, thereby conserving computational resources.
Several mechanisms for constructing the gating function $ G(x) $ exist ~\citep{clark2022unified,hazimeh2021dselect,zhou2022mixture};
however, our implementation opts for the efficient approach of selecting the Top-K values from a linear projection, followed by a softmax operation~\citep{shazeer2017outrageously}:
\vspace{-5pt}
\begin{equation}
\centering
G(x) = Softmax(TopK(xW_g)) 
\end{equation}

in which $ TopK $ preserves the highest K values from the logit vector $ \ell \in \mathbb{R}^n $, assigning them as $-\infty$ otherwise. 
The choice of K, indicative of the active experts per token, is a strategic hyperparameter influencing the computational expenditure per token.
Elevating the number of experts $ n $ while maintaining a constant K allows for an augmentation of the model's capacity without a proportionate rise in computational overhead.
This strategy creates a distinction between the model's total parameters or the sparse parameter count, and the activeparameter count, which relates directly to K and determines the number of parameters actively employed per token.

\begin{figure}[htbp]
\centering
\includegraphics[width=\linewidth]{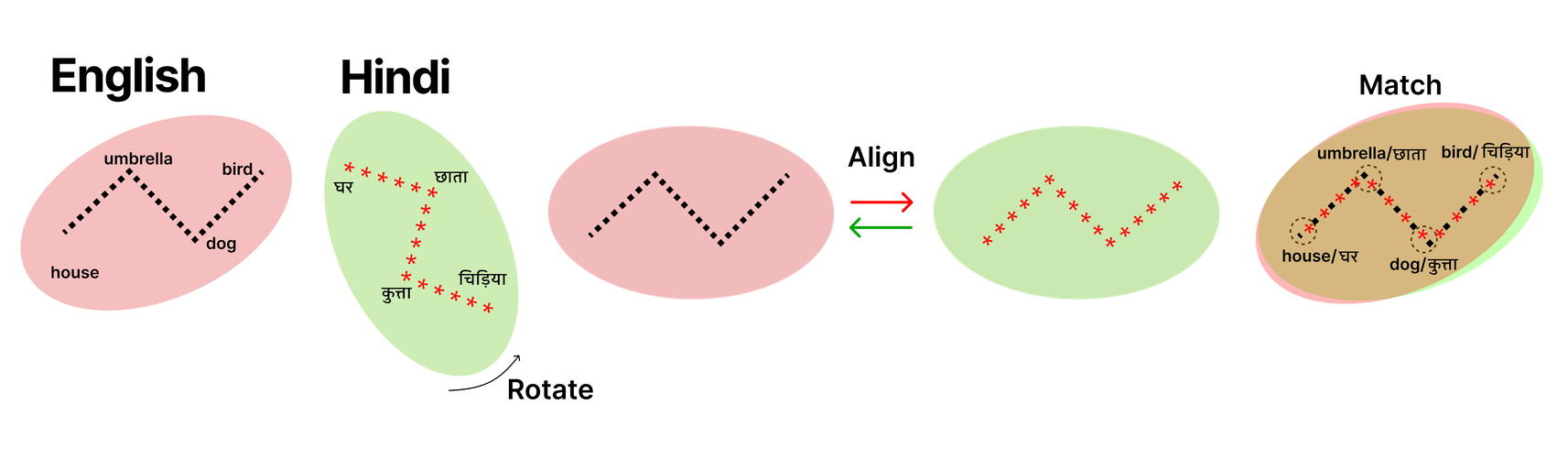}
  \caption{The same concepts (umbrella, house, dog) when expressed in different languages (English, Hindi) can be mapped to quite different embedding vectors (left). In order to achieve multilingual encoders and decoders which map concepts in different languages to a common concept space, the these embedding vectors need to be aligned (middle). It can be seen that after the multilingual concept alignment stage, the same concepts (umbrella, house, dog) are now mapped to similar embedding vectors, even though they are expressed in different languages (right). Our specialized Neural Machine Translation (NMT) based encoders and decoders, can apply the same principle to master multi-language translation and ensure concept consistency across languages.}
\label{fig:language_matching}
\end{figure}

\subsection{Training Data}

The key to our language training strategy lies in leveraging linguistic commonalities during the language learning phase.
For example, Hindi has a lot more commonalities in terms of semantics, grammar and cultural context with Gujarati or Bengali as compared to Nordic languages.

The limited multilingual capabilities of large language models (LLMs) stem from an uneven data distribution favoring a handful of well-resourced languages.
Multilingual data in machine translation is task-specific and misses key training areas for LLMs, such as conversation, summarization, and instruction-following.
To address this, the SUTRA dataset includes over 100 million conversations in real and synthetically translated pairs across various languages, supplemented by publicly available datasets for a comprehensive training landscape.
Past research has demonstrated synthetic data's role in fostering reasoning, code generation, and task complexity learning in LLMs, as well as enhancing cross-lingual transfer with multilingual synthetic data \citep{lai2023okapi,whitehouse2023llm}.
Following this insight, we adopt a methodical use of abundant data from languages like English to facilitate concept learning.
During the language learning and alignment phases of multilingual training, we employ a combination of real and synthetic data to bolster and broaden our training framework.
An illustraion and description of the matching and alignment process is shown in Figure~\ref{fig:language_matching}.

\begin{figure}[tbp]
\centering
\includegraphics[width=\linewidth]{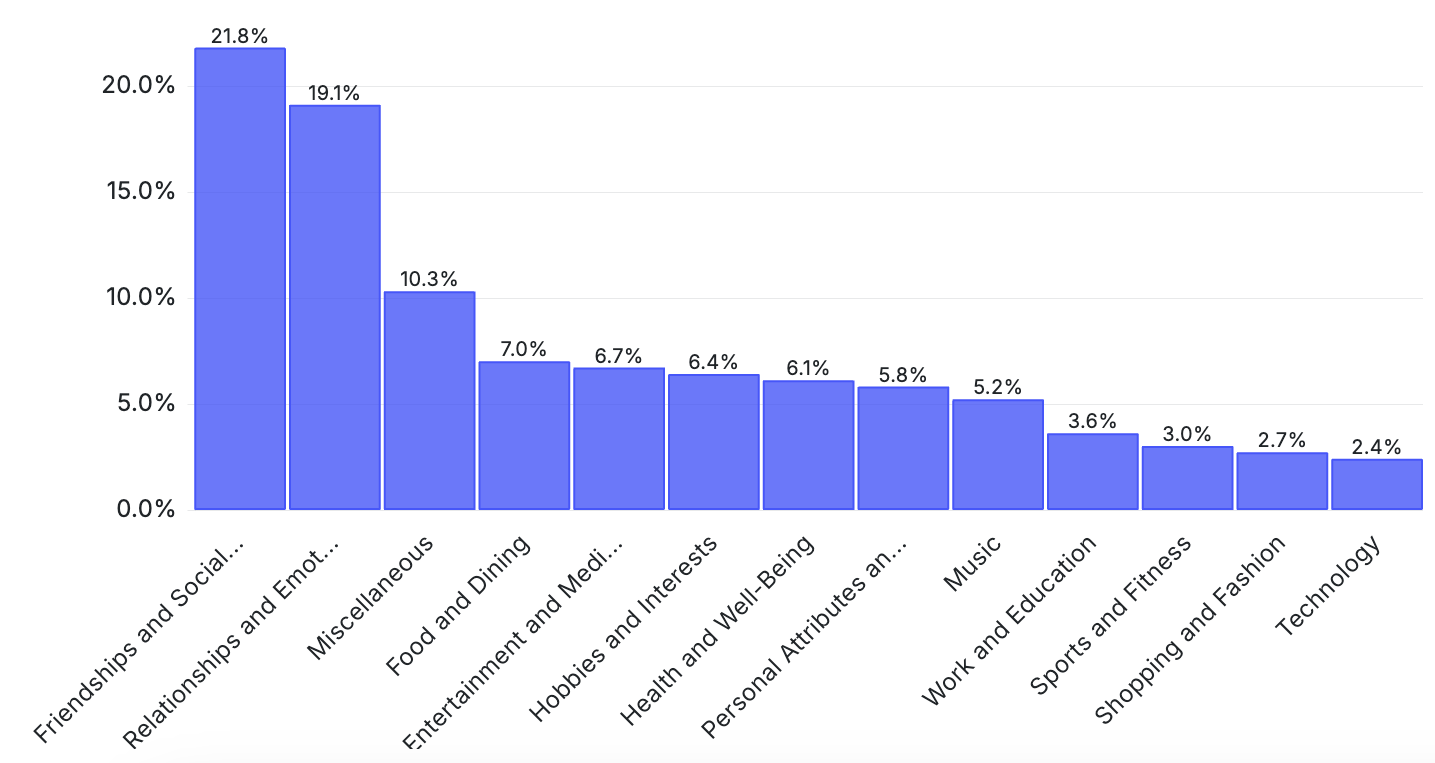}
  \caption{\textbf{Conversation Data Topic Distribution}. In the following plot we are are showing topic distribution of over 1M sampled conversations. Inspection of cluster centroids reveals that this is a rich and diverse data covering wide range of topics.}
\label{fig:conversation_data_distribution}
\end{figure}

In Figure~\ref{fig:conversation_data_distribution} we show the topic distribution of over 1M sampled conversations.
Inspection of cluster centroids reveals that this is a rich and diverse data covering a wide range of topics. 


The purpose-built multi-language tokenizers efficiently represent each language. They are trained on cross-lingual data and finely tuned with the base model, setting a new benchmark in multilingual language modeling. 
One of the most critical aspects of having good performance of conversational LLMs is high-quality instruction fine-tuning (IFT) datasets.
Majority of IFT datasets are in English. We use Neutral Machine Translation (NMT) to translate the instructions, inputs and outputs from different datasets, to ensure balanced representation across tasks, in multiple Indian and non-English languages.
Overall, we prepare more than 100M training samples from languages like English, Hindi, Tamil, Korean etc. with wide ranging datasets such as our internal SUTRA dataset, as well as open-source FLAN-v2, OpenAssistant and wikiHow. 
The translated examples are filtered to retain high-quality examples.
Note that our internal data includes long-term and multi-turn conversational data, which helps to tune it towards better human-AI conversations and interactions.
A comparison and detailed description of the dataset is shown in Table~\ref{table:datasets_summary}.

\begin{table}[tb]
\setlength{\abovecaptionskip}{10pt} 
\setlength{\belowcaptionskip}{5pt} 
\centering
  \begin{tabular}{lrrrr}
  \toprule
  \textbf{Dataset}          & \textbf{Number} & \textbf{\# Users} & \textbf{Avg. Turns / Sample} & \textbf{Avg. Tokens / Prompt} \\ \midrule
  Anthropic HH              & 338,704         & 143               & 2.3                         & 19                            \\
  OpenAssistant             & 66,497          & 13,500            & -                           & 37                            \\
  Chatbot Arena             & 33,000          & 13,383            & 2.1                         & 53                            \\
  SUTRA dataset             & 10M+            & 350,000           & 15                          & 120                           \\
  \bottomrule
  \end{tabular}

\caption{In this table, statistics of various leading conversation datasets are shown such as Anthropic HH \citep{bai2022training}, OpenAssistant Conversations \citep{kopf2023openassistant}, LMSys \citep{chiang2024chatbot} and the SUTRA dataset. The tokens are counted using Llama2 tokenizer \citep{touvron2023llama2} for public datasets and for SUTRA dataset using SUTRA's tokenizer. One of the key aspects of our dataset is having long term and multi-turn conversations.}

\label{table:datasets_summary}
\end{table}


\section{Training Multilingual Tokenizers}
Tokenization, a critical step in NLP pipeline, involves converting text into a sequence of tokens, where each token represents a subword or word. Although English specific tokenizers can generate text in non-English languages, they don't capture language specific nuances and are highly inefficient in other languages, especially non-Romanized languages. More specifically for Indian languages like Hindi, Gujarati, or Tamil, we note that tokenizers from leading LLMs like Llama-2, Mistral, and GPT-4 consume 4.5X to 8X more tokens compared to English, as shown in Table~\ref{tab:token_counts}.


\begin{table}[ht]
\setlength{\abovecaptionskip}{10pt} 
\setlength{\belowcaptionskip}{5pt} 
\centering
\begin{tabular}{
  >{\raggedright\arraybackslash}p{1.5cm}
  >{\centering\arraybackslash}p{2.5cm}
  >{\centering\arraybackslash}p{1.9cm}
  >{\centering\arraybackslash}p{1.7cm}
  >{\centering\arraybackslash}p{3.3cm}
  >{\centering\arraybackslash}p{2.0cm}
}
\toprule
\textbf{Model/Tok.} & \textbf{English} & \textbf{Hindi} & \textbf{Gujarati} & \textbf{Tamil} & \textbf{Korean} \\
\midrule
  & {\includegraphics[width=25mm, height=16mm]{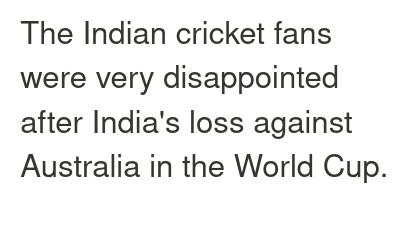}}
  & {\includegraphics[width=19mm, height=16mm]{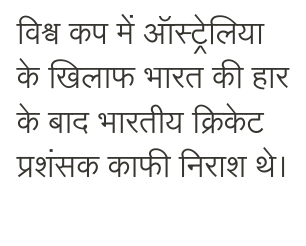}}
  & {\includegraphics[width=17mm, height=16mm]{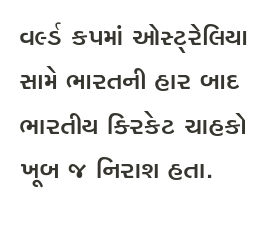}}
  & {\includegraphics[width=33mm, height=16mm]{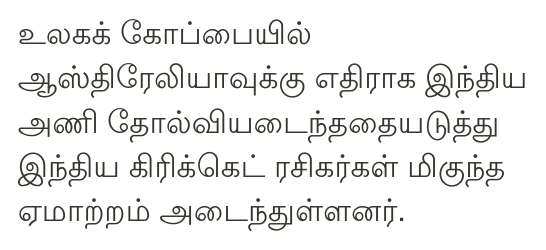}}
  & {\includegraphics[width=20mm, height=16mm]{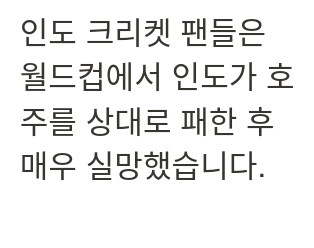}}
  \\
\midrule
SUTRA   & 19 & \textbf{18} & \textbf{23} & \textbf{22} & \textbf{20} \\
GPT-4   & \textbf{18} & 82 & 137 & 180 & 100 \\
Llama2  & 22 & 98 & 200 & 163 & 68 \\
Mistral & 19 & 93 & 165 & 157 & 53 \\
\bottomrule
\end{tabular}
\caption{Number of tokens per language for models (Note: Lower is better). SUTRA Models use these multilingual tokenizers to get quality, performance, and efficiency.}
\label{tab:token_counts}
\end{table}

A key step in adding language specific skills is decreasing the average number of tokens a word is split into (also known as token fertility) by a language model on non-english text. This makes inferencing efficient as well as semantically meaningful. We train the sentence-piece tokenizer from a large corpus of multi-language dataset of 500K+ documents, which is then merged with a pre-trained english tokenizer to increase the vocabulary size. Text generated with our tokenizers lead to 80\% to 200\% reduction in overall tokens consumed across languages, which is critical for bringing down the cost of inferencing when deploying these models for cost-sensitive use-cases.

\section{Multilingual MMLU}

\subsection{Massive Multitask Language Understanding}
We evaluate our model on a variety of NLU and NLG tasks. To test the knowledge and reasoning capabilities of the model, we evaluate on the machine-translated version of the benchmarks such as MMLU \citep{hendrycks2021measuring}. The Massive Multitask Language Understanding (MMLU) benchmark is a comprehensive and challenging evaluation framework designed to test the capabilities of Large Language Models (LLMs) across a wide array of tasks. It was created with the goal of pushing the boundaries of what LLMs can understand and how well they can adapt to various domains of knowledge. The benchmark covers 57 subjects across STEM, the humanities, the social sciences, and more. It ranges in difficulty from an elementary level to an advanced professional level, and it tests both world knowledge and problem solving ability. Subjects range from traditional areas, such as mathematics and history, to more specialized areas like law and ethics. The granularity and breadth of the subjects makes the benchmark ideal for identifying a model’s blind spots. This diversity ensures that models are not only proficient in a broad spectrum of topics but also capable of generalizing their understanding to new and unseen domains. The MMLU evaluates models on their ability to answer multiple-choice questions, requiring nuanced comprehension and the application of reasoning, which collectively serve as a measure of an LLM's depth of knowledge and its interpretive skills.


\subsection{Extending MMLU to Multiple Languages}
To assess our models' effectiveness in various tasks and across multiple languages, we developed a multilingual evaluation suite that broadens the scope of evaluation linguistically. We utilized the multilingual assessment framework suggested by \citet{lai2023okapi} and \citet{ustun2024aya}, with certain distinctions. Notably, while Okapi uses a 25-shot evaluation, our methodology employs a 5-shot evaluation as per the original benchmark by \citet{hendrycks2021measuring}. We anticipate that a 5-shot evaluation, offering fewer examples, presents a more challenging benchmark. Recognizing the existence of over 200 major languages globally, our evaluation focuses on three distinct language groups: English, Korean, Japanese, Arabic, and Indian Languages. Although this selection is not exhaustive, it encompasses a significant portion of linguistic diversity, enabling thorough analysis of the models' multilingual capabilities. These languages represent a substantial demographic, accounting for more than half of the global population as primary or secondary speakers. Additionally, they are key languages in global business, ensuring our evaluation has broad relevance.

\subsection{Consistent Performance across Languages}

The SUTRA model demonstrates a notable consistency in linguistic performance across a variety of languages, as evidenced by the MMLU benchmark results. It exhibits a minimal performance deviation from its English language results to other languages such as Hindi, Gujarati, and Arabic, highlighting its robust multilingual capabilities critical for applications on a global scale.

Superior concept and language modeling underpin the SUTRA model's ability to maintain performance levels across different languages, distinguishing it from other leading models, including GPT-4, GPT-3.5, and Llama2.
Many existing model architectures (including purpose built multilanguage models) experience a pronounced decline in performance in non-English languages, often regressing to baseline random chance performance, as detailed in Table~\ref{table:purposebuilt_mmlu}. Note that random chance performance is at 25\% on the MMLU benchmark. In contrast, SUTRA consistently achieves stable scores across languages, setting it apart, particularly in languages that are less commonly represented in language models, such as Hindi, Gujarati, Tamil, and Korean.
The SUTRA model, therefore, not only excels in individual language performance but also promotes a more universal, language-agnostic approach to AI.
It serves as a robust solution for international businesses, educational platforms, and cross-cultural communication, setting a new benchmark for LLMs in a multi-lingual, interconnected world.

\begin{table}[tb]
\setlength{\abovecaptionskip}{10pt} 
\setlength{\belowcaptionskip}{5pt} 
\centering

    \begin{tabular}{lcccccccccccc}
        \toprule
        \textbf{Model Name} & \textbf{en} & \textbf{hi} & \textbf{gu} & \textbf{ta} & \textbf{ko} & \textbf{bn} & \textbf{pa} & \textbf{mr} & \textbf{te} & \textbf{ml} & \textbf{ar} & \textbf{ja} \\
        \midrule
        SUTRA & 77 & 68 & 67 & 67 & 67 & 68 & 67 & 69 & 68 & 68 & 67 & 75 \\
                Okapi & - & 26 & 27 & 26 & - & 26 & - & 26 & 26 & 25 & 27 & - \\
        mT0 & - & 32 & 29 & 29 & - & 31 & - & 31 & 29 & 29 & 31 & - \\
        mT0X & - & 31 & 29 & 27 & - & 30 & - & 29 & 27 & 27 & 31 & - \\
        Aya & - & 39 & 33 & 21 & - & 36 & - & 36 & 39 & 32 & 38 & - \\
        \bottomrule
    \end{tabular}

  \caption{
 The above table shows comparison with recent purpose built multilingual language models such as those proposed by \citet{ustun2024aya}, \citet{lai2023okapi}. SUTRA provides strong multilingual performance compared to many leading purpose built multilingual language models by significant margin.  }

\label{table:purposebuilt_mmlu}
\end{table}


\subsection{Comparing with leading models for Multilingual Performance}


\begin{table}[h]
\setlength{\abovecaptionskip}{10pt} 
\setlength{\belowcaptionskip}{5pt} 
\centering
\begin{tabular}{@{}lcccccccccc@{}}
\toprule
\textbf{Language} & \textbf{SUTRA} & \textbf{LL3 70B} & \textbf{LL2 70B} & \textbf{GPT 4} & \textbf{GPT 3.5} & \textbf{Mixt. 8x22B} & \textbf{Mixt. 8x7B} & \textbf{HCX} & \textbf{PPLX} \\ \midrule
English         & 77    & 82          & 63           & 86    & 70       & 77            & 70          & 66          & 62                        \\
Hindi           & 68    & 64          & 31           & 71    & 39       & 38            & 35          & 39          & 32                        \\
Korean          & 67    & 60          & 38           & 72    & 51       & 56            & 46          & 54          & 40                        \\
Gujarati        & 67    & 54          & 29           & 61    & 35       & 29            & 29          & 36          & 26                        \\
Tamil           & 67    & 52          & 29           & 44    & 30       & 34            & 29          & 33          & 27                        \\
Bengali         & 68    & 58          & 27           & 73    & 36       & 37            & 33          & --          & --                        \\
Punjabi         & 67    & 55         & 26           & 71    & 34       & 29            & 30          & --          & --                        \\
Marathi         & 69    & 62          & 25           & 66    & 32       & 36            & 32          & --          & --                        \\
Telugu          & 68    & 53          & 24           & 62    & 32       & 32            & 28          & --          & --                        \\
Arabic          & 67    & 60          & 48           & 80    & 49       & 48            & 39          & --          & --                        \\
Japanese        & 75    & 70          & 56           & 80    & 57       & 60            & 51          & --          & --                        \\ \bottomrule
\end{tabular}
\caption{The table shows multilingual performance of various leading models on MMLU benchmark for multiple languages. SUTRA has competitive performance in English while maintaining strong multilingual performance in other languages. Many leading language models' MMLU scores for non-english languages falls close to random chance (25\% is random chance on MMLU task).}
\label{table:leading_mmlu1}
\end{table}

For our evaluation, we use multiple state of the art models and compare their performance on the multilingual MMLU benchmark, as shown in Table~\ref{table:leading_mmlu1}. We considered multiple leading models such as GPT-4 and GPT-3.5 from OpenAI, Mixtral-8x7b from Mistral, Llama2-13b, Llama2-70b and Llama3-70b from Meta,  sonar-medium from Perplexity, HyperClovaX from Naver, and Airavata Model from Sarvam AI. Of these, GPT-4, GPT-3.5, Mixtral, Llama series and Perplexity are generic models i.e. they were not trained to optimize for specific languages. HyperClovaX was specifically trained to optimize performance on the Korean language, whilst Airavata was specifically trained to optimize performance in Hindi.

Overall, the evaluation results demonstrate that our SUTRA models can match and even outperform GPT-3.5 and Llama-7b on TWO-related use cases, particularly for providing natural and engaging responses across languages. Although GPT-4 is still state-of-the-art in terms of performance, cost continues to be a major hindrance for wide-scale deployment in cost-sensitive markets. Surpassing GPT-3.5 multilingual performance by 20-30\% on the leading MMLU benchmark, SUTRA models excel in comprehending and generating responses across numerous languages. We find that SUTRA does well even compared to models that were specifically optimized for a particular language, showing promise for the approach followed by SUTRA, as shown in Table~\ref{table:language_specific_mmlu}.
More detailed results showing MMLU scores across groups of categories such as STEM, humanities etc. are listed in Table~\ref{table:quantitative_mmlu_results_sutra_pro}.

\begin{table}[h]
\setlength{\abovecaptionskip}{10pt} 
\setlength{\belowcaptionskip}{5pt} 
\centering
\begin{tabular}{lllll}
\toprule
\textbf{Language} & \textbf{Organization} & \textbf{Model Name} &  \textbf{MMLU} & \textbf{SUTRA MMLU} \\ \midrule
Hindi    & Sarvam       & Airavata   & 35            & \textbf{68}         \\
Korean   & Naver        & HyperClovaX & 54           & \textbf{67}         \\
Arabic   & Inception / MBZUAI & Jais & 34            & \textbf{67}         \\
Japanese & Rakuten      & Rakuten-7B & 61            & \textbf{75}         \\ \bottomrule
\end{tabular}
\caption{The above table shows comparison of language specific LLMs for multiple languages such as Hindi \citep{gala2024airavata}, Korean \citep{son2024kmmlu}, Arabic \citep{sengupta2023jais} and Japanese \citep{group2024rakutenai}. The selected models are best performing models on respective languages, as they were purposely built and tuned for those languages. Shown on the right is MMLU score for SUTRA on respective languages. From the performance numbers it is evident that the concept and language modeling approach followed by SUTRA yields superior multilingual performance.}
\label{table:language_specific_mmlu}
\end{table}

\begin{table}[tb]
\setlength{\abovecaptionskip}{10pt} 
\setlength{\belowcaptionskip}{5pt} 
\centering

  \begin{tabular}{lclccccc}
  \toprule
    \textbf{Language}  & \textbf{STEM} & \textbf{Social Sci.} & \textbf{Humanities} & \textbf{Other} & \textbf{$\sim$Average} \\
  \midrule
  English  & 69.78 & 80.83 & 76.08 & 79.07 & 76.24 \\
  Hindi    & 61.67 & 74.17 & 69.23 & 68.93 & 67.81 \\
  Gujarati &  62.78 & 70.83 & 66.92 & 66.79 & 66.40 \\
  Marathi  &  59.72 & 75.42 & 73.46 & 71.07 & 68.95 \\
  Bengali  &  60.83 & 74.17 & 66.54 & 73.57 & 68.07 \\
  Tamil    &  61.39 & 71.67 & 64.62 & 70.71 & 66.58 \\
  Punjabi  &  59.72 & 73.75 & 69.62 & 68.21 & 67.02 \\
  Korean   & 57.78 & 72.08 & 67.69 & 70.30 & 66.14 \\
  Arabic   & 59.72 & 75.83 & 63.85 & 68.93 & 66.32 \\
  Japanese & 66.67 & 81.67 & 76.15 & 78.57 & 74.91 \\
  \bottomrule
  \end{tabular}

  \caption{
    SUTRA quantitative MMLU results across a subset of supported languages for fine-grained tasks on the MMLU benchmark.
  }

\label{table:quantitative_mmlu_results_sutra_pro}
\end{table}

%

\section{Quantitative Evaluation for Real-Time Queries}

SUTRA models are connected, up-to-date, and hallucination-free models that provide factual responses with a conversational tone. They are online LLMs that use, infer, and process real-time knowledge from the internet and leverage it to provide the most up-to-date information when forming responses. SUTRA-Online models can accurately respond to time-sensitive queries, extending its knowledge beyond a static training corpus. Online models can therefore accurately answer questions like "Who won the game last night” or “What’s the most popular movie right now?”.

\begin{table}[tb]
\setlength{\abovecaptionskip}{10pt} 
\setlength{\belowcaptionskip}{5pt} 
\centering
\begin{tabular}{@{}llllllllll@{}}
\toprule
\textbf{Model Name} & \textbf{Knowledge Cut.} & \textbf{FreshLLM Dataset} & \textbf{all} & \textbf{all} & \textbf{fast} & \textbf{slow} & \textbf{never} & \textbf{$\geq$ 2022} & \textbf{1-hop} \\ 
\midrule
Google Search & Up to Date & 04/26/2023 & 39.6 & 48.9 & 32 & 46.4 & 68.3 & 37.9 & 55.6 \\
GPT-3.5 & 2021 & 04/26/2023 & 26 & 26.1 & 4 & 15.2 & 58.7 & 5.1 & 28 \\
Google DeepMind & Up to Date & 04/26/2023 & 56 & 62.5 & 46.4 & 60.8 & 80.2 & 57 & 68.7 \\
Perplexity AI & Up to Date & 04/26/2023 & 52.2 & 57.2 & 38.4 & 53.6 & 79.4 & 47.7 & 63.8 \\
SUTRA-Online & Up to Date & 04/15/2024 & \textbf{56} & \textbf{63.8} & \textbf{47.7} & \textbf{61.6} & \textbf{88.7} & \textbf{59.1} & \textbf{70.4} \\
\bottomrule
\end{tabular}
\caption{Performance Comparison of Language Models for handling fresh (realtime queries) with valid premise according to freshness LLM benchmark from \citet{vu2023freshllms}}
\label{table:freshness_results}
\end{table}

We evaluated the SUTRA models using the Fresh Prompt framework \citep{vu2023freshllms}, developed by Google for assessing online LLMs \citep{press2022measuring}, and discovered that SUTRA-Online models surpass the competing search engine-augmented models from Google, as well as OpenAI's GPT-3.5 and Perplexity AI. The benchmark contains exhaustive questions covering various nuanced online scenarios covering never-changing, in which the answer almost never changes; slow-changing, in which the answer typically changes over the course of several years; fast-changing, in which the answer typically changes within a year or less. SUTRA performed well across majority of these scenarios, as shown in Table~\ref{table:freshness_results}.

\section{Discussion and Conclusion}

Looking ahead, the SUTRA paves the way for the development of phonetic models (approach for SUTRA-Dhvanim), which benefits from the clear separation between concept modeling and language learning. By replacing the NMT decoder with a phonetic decoder, we enable the generation of phonetic responses for more seamless integration with speech models. Our next frontier for optimization is to examine the accuracy and performance impact of structured sparsity and int4 precision, which could significantly reduce SUTRA's GPU memory footprint and yield with improvements in latency. 

This research has introduced SUTRA, a state-of-the-art multilingual conversational language model, showcasing its superior ability to handle multiple languages with remarkable efficiency and performance.
SUTRA is already proficient in 31 languages across multiple tasks, as detailed in Table~\ref{table:supported_languages}, and is being extended to support over 50 languages.
Unlike its predecessors, which struggle with the nuanced requirements of multi-language understanding, SUTRA exhibits a robust proficiency that is evident across a range of linguistic contexts. This is particularly notable in its application to languages with fewer resources available for model training, which traditionally lag in performance metrics. The innovative architecture of SUTRA, with its decoupled concept and language processing, allows for a scalable and flexible approach to language model training. This not only opens the door for more equitable representation of less commonly spoken languages but also ensures that the quality of interaction remains high across all languages. The efficient tokenization strategy of SUTRA, reducing token fertility for non-English languages, also points to potential cost reductions in deploying AI in multi-language environments, a notable consideration for global accessibility.

In conclusion, SUTRA sets a new precedent for multilingual language models by delivering high performance and efficiency without sacrificing linguistic diversity. Its architecture, which mirrors human cognitive development by separating concept understanding from linguistic expression, allows for a more natural and extensive language comprehension. This breakthrough bears significant implications for the global adoption and application of AI, paving the way for more inclusive and equitable access to technology across language barriers.

\begin{table}[tb]
\setlength{\abovecaptionskip}{10pt} 
\setlength{\belowcaptionskip}{5pt} 
\centering
\begin{tabular}{llll}
\toprule
\textbf{Language}  & \textbf{ISO Code} & \textbf{Language}    & \textbf{ISO Code} \\ \midrule
English            & en                & Korean               & ko                \\
French             & fr                & Japanese            & ja                \\
Italian            & it                & Thai                & th                \\
Spanish            & es                & Arabic              & ar                \\
German             & de                & Persian             & fa                \\
Portuguese         & pt                & Vietnamese          & vi                \\
Hindi              & hi                & Indonesian          & id                \\
Bengali            & bn                & Turkish             & tr                \\
Marathi            & mr                & Polish              & pl                \\
Telugu             & te                & Russian             & ru                \\
Tamil              & ta                & Ukranian            & uk                \\
Gujarati           & gu                & Dutch               & nl                \\
Kannada            & kn                & Greek               & el                \\
Malayalam          & ml                &                     &                   \\
Punjabi            & pa                &                     &                   \\
Assamese           & as                &                     &                   \\
Urdu               & ur                &                     &                   \\
Odia               & or                &                     &                   \\ \bottomrule
\end{tabular}
\caption{Although SUTRA can support more than 50 languages, the languages listed in the table above are the ones we have tested across number of tasks. Support for additional languages will be released in next versions of SUTRA}
\label{table:supported_languages}
\end{table}

\bibliographystyle{unsrtnat}
\bibliography{references}

\begin{thebibliography}{38}
\providecommand{\natexlab}[1]{#1}
\providecommand{\url}[1]{\texttt{#1}}
\expandafter\ifx\csname urlstyle\endcsname\relax
  \providecommand{\doi}[1]{doi: #1}\else
  \providecommand{\doi}{doi: \begingroup \urlstyle{rm}\Url}\fi

\bibitem[Brown et~al.(2020)]{brown2020language}
Tom~B. Brown et~al.
\newblock Language models are few-shot learners.
\newblock \emph{arXiv preprint arXiv:2005.14165}, 2020.

\bibitem[Devlin et~al.(2018)Devlin, Chang, Lee, and Toutanova]{devlin2018bert}
Jacob Devlin, Ming-Wei Chang, Kenton Lee, and Kristina Toutanova.
\newblock Bert: Pre-training of deep bidirectional transformers for language
  understanding.
\newblock \emph{arXiv preprint arXiv:1810.04805}, 2018.

\bibitem[Jia et~al.(2019)]{jia2019bias}
Roger Jia et~al.
\newblock Bias in multilingual models: The case for linguistic equity in ai.
\newblock In \emph{Conference on Neural Information Processing Systems
  (NeurIPS)}, 2019.

\bibitem[Conneau et~al.(2020)]{conneau2020unsupervised}
Alexis Conneau et~al.
\newblock Unsupervised cross-lingual representation learning at scale.
\newblock \emph{arXiv preprint arXiv:1911.02116}, 2020.

\bibitem[Smith et~al.(2021)]{smith2021can}
Linda Smith et~al.
\newblock Can multilingual models transfer for less resourced languages?
\newblock \emph{Language Resources and Evaluation}, 2021.

\bibitem[Zhang et~al.(2020)]{zhang2020improving}
Yiming Zhang et~al.
\newblock Improving multilingual models with language-clustered vocabularies.
\newblock \emph{arXiv preprint arXiv:2007.07680}, 2020.

\bibitem[Wu et~al.(2019)]{wu2019google}
Yonghui Wu et~al.
\newblock Google’s neural machine translation system: Bridging the gap
  between human and machine translation.
\newblock \emph{arXiv preprint arXiv:1609.08144}, 2019.

\bibitem[Shazeer et~al.(2017)]{shazeer2017outrageously}
Noam Shazeer et~al.
\newblock Outrageously large neural networks: The sparsely-gated
  mixture-of-experts layer.
\newblock \emph{arXiv preprint arXiv:1701.06538}, 2017.

\bibitem[Hendrycks et~al.(2021)]{hendrycks2021measuring}
Dan Hendrycks et~al.
\newblock Measuring massive multitask language understanding.
\newblock \emph{arXiv preprint arXiv:2009.03300}, 2021.

\bibitem[Koehn and Knowles(2017)]{koehn2017six}
Philipp Koehn and Rebecca Knowles.
\newblock Six challenges for neural machine translation.
\newblock \emph{arXiv preprint arXiv:1706.03872}, 2017.

\bibitem[Son and Kim(2023)]{son2023translation}
Jungwoo Son and Byeongil Kim.
\newblock Translation performance from the user's perspective of large language
  models and neural machine translation systems.
\newblock \emph{Information}, 14\penalty0 (10):\penalty0 574, 2023.

\bibitem[Wu et~al.(2016)Wu, Schuster, Chen, Le, Norouzi, Macherey, Krikun, Cao,
  Gao, Macherey, et~al.]{wu2016google}
Yonghui Wu, Mike Schuster, Zhifeng Chen, Quoc~V Le, Mohammad Norouzi, Wolfgang
  Macherey, Maxim Krikun, Yuan Cao, Qin Gao, Klaus Macherey, et~al.
\newblock Google's neural machine translation system: Bridging the gap between
  human and machine translation.
\newblock \emph{arXiv preprint arXiv:1609.08144}, 2016.

\bibitem[Costa-juss{\`a} et~al.(2022)Costa-juss{\`a}, Cross, {\c{C}}elebi,
  Elbayad, Heafield, Heffernan, Kalbassi, Lam, Licht, Maillard,
  et~al.]{costa2022no}
Marta~R Costa-juss{\`a}, James Cross, Onur {\c{C}}elebi, Maha Elbayad, Kenneth
  Heafield, Kevin Heffernan, Elahe Kalbassi, Janice Lam, Daniel Licht, Jean
  Maillard, et~al.
\newblock No language left behind: Scaling human-centered machine translation.
\newblock \emph{arXiv preprint arXiv:2207.04672}, 2022.

\bibitem[Zhou et~al.(2022)Zhou, Lei, Liu, Du, Huang, Zhao, Dai, Le, Laudon,
  et~al.]{zhou2022mixture}
Yanqi Zhou, Tao Lei, Hanxiao Liu, Nan Du, Yanping Huang, Vincent Zhao, Andrew~M
  Dai, Quoc~V Le, James Laudon, et~al.
\newblock Mixture-of-experts with expert choice routing.
\newblock \emph{Advances in Neural Information Processing Systems},
  35:\penalty0 7103--7114, 2022.

\bibitem[Zoph(2022)]{zoph2022designing}
Barret Zoph.
\newblock Designing effective sparse expert models.
\newblock \emph{IEEE International Parallel and Distributed Processing
  Symposium (IPDPS)}, 2022.

\bibitem[Ott et~al.(2022)Ott, Shleifer, Shuster, Simig, Koura, Sridhar, Wang,
  and Zettlemoyer]{ott2022opt}
Myle Ott, Sam Shleifer, Kurt Shuster, Daniel Simig, P~Sai Koura, Abhinav
  Sridhar, Tao Wang, and Luke Zettlemoyer.
\newblock Opt: Open pre-trained transformer language models.
\newblock 2022.

\bibitem[Zheng et~al.(2019)Zheng, Huang, and Sun]{zheng2019chid}
Chujie Zheng, Minlie Huang, and Aixin Sun.
\newblock Chid: A large-scale chinese idiom dataset for cloze test.
\newblock In \emph{Proceedings of the 57th Annual Meeting of the Association
  for Computational Linguistics}, pages 778--787, 2019.
\newblock \doi{10.18653/v1/P19-1075}.
\newblock URL \url{https://doi.org/10.18653/v1/p19-1075}.

\bibitem[Dai et~al.(2019)Dai, Yang, Yang, Carbonell, Le, and
  Salakhutdinov]{dai2019transformer}
Zihang Dai, Zhilin Yang, Yiming Yang, Jaime~G Carbonell, Quoc Le, and Ruslan
  Salakhutdinov.
\newblock Transformer-xl: Attentive language models beyond a fixed-length
  context.
\newblock \emph{arXiv preprint arXiv:1901.02860}, 2019.

\bibitem[Nichols and Warnow(2008)]{nichols2008tutorial}
Johanna Nichols and Tandy Warnow.
\newblock Tutorial on computational linguistic phylogeny.
\newblock \emph{Language and Linguistics Compass}, 2\penalty0 (5):\penalty0
  760--820, 2008.

\bibitem[Birch(2021)]{birch2021neural}
Alexandra Birch.
\newblock \emph{Neural Machine Translation}.
\newblock Cambridge University Press, 2021.

\bibitem[Vu et~al.(2023)Vu, Iyyer, Wang, Constant, Wei, Wei, Tar, Sung, Zhou,
  Le, and Luong]{vu2023freshllms}
Tu~Vu, Mohit Iyyer, Xuezhi Wang, Noah Constant, Jerry Wei, Jason Wei, Chris
  Tar, Yun-Hsuan Sung, Denny Zhou, Quoc Le, and Thang Luong.
\newblock Freshllms: Refreshing large language models with search engine
  augmentation, 2023.

\bibitem[Press et~al.(2022)Press, Zhang, Min, Schmidt, Smith, and
  Lewis]{press2022measuring}
Ofir Press, Muru Zhang, Sewon Min, Ludwig Schmidt, Noah~A Smith, and Mike
  Lewis.
\newblock Measuring and narrowing the compositionality gap in language models.
\newblock \emph{arXiv preprint arXiv:2210.03350}, 2022.

\bibitem[Komeili et~al.(2021)Komeili, Shuster, and Weston]{komeili2021internet}
Mojtaba Komeili, Kurt Shuster, and Jason Weston.
\newblock Internet-augmented dialogue generation.
\newblock \emph{arXiv preprint arXiv:2107.07566}, 2021.

\bibitem[Vaswani et~al.(2017)]{vaswani2017attention}
Ashish Vaswani et~al.
\newblock Attention is all you need.
\newblock \emph{Advances in neural information processing systems}, 30, 2017.

\bibitem[Jiang et~al.(2023)Jiang, Sablayrolles, Mensch, Bamford, Chaplot,
  Casas, Bressand, Lengyel, Lample, Saulnier, et~al.]{jiang2023mistral}
Albert~Q Jiang, Alexandre Sablayrolles, Arthur Mensch, Chris Bamford,
  Devendra~Singh Chaplot, Diego de~las Casas, Florian Bressand, Gianna Lengyel,
  Guillaume Lample, Lucile Saulnier, et~al.
\newblock Mistral 7b.
\newblock \emph{arXiv preprint arXiv:2310.06825}, 2023.

\bibitem[Clark et~al.(2022)Clark, De~Las~Casas, Guy, Mensch, Paganini,
  Hoffmann, Damoc, Hechtman, Cai, Borgeaud, et~al.]{clark2022unified}
Aidan Clark, Diego De~Las~Casas, Aurelia Guy, Arthur Mensch, Michela Paganini,
  Jordan Hoffmann, Bogdan Damoc, Blake Hechtman, Trevor Cai, Sebastian
  Borgeaud, et~al.
\newblock Unified scaling laws for routed language models.
\newblock In \emph{International Conference on Machine Learning}, pages
  4057--4086. PMLR, 2022.

\bibitem[Hazimeh et~al.(2021)Hazimeh, Zhao, Chowdhery, Sathiamoorthy, Chen,
  Mazumder, Hong, and Chi]{hazimeh2021dselect}
Hussein Hazimeh, Zhe Zhao, Aakanksha Chowdhery, Maheswaran Sathiamoorthy, Yihua
  Chen, Rahul Mazumder, Lichan Hong, and Ed~Chi.
\newblock Dselect-k: Differentiable selection in the mixture of experts with
  applications to multi-task learning.
\newblock \emph{Advances in Neural Information Processing Systems},
  34:\penalty0 29335--29347, 2021.

\bibitem[Lai et~al.(2023)Lai, Van~Nguyen, Ngo, Nguyen, Dernoncourt, Rossi, and
  Nguyen]{lai2023okapi}
Viet~Dac Lai, Chien Van~Nguyen, Nghia~Trung Ngo, Thuat Nguyen, Franck
  Dernoncourt, Ryan~A Rossi, and Thien~Huu Nguyen.
\newblock Okapi: Instruction-tuned large language models in multiple languages
  with reinforcement learning from human feedback.
\newblock \emph{arXiv preprint arXiv:2307.16039}, 2023.

\bibitem[Whitehouse et~al.(2023)Whitehouse, Choudhury, and
  Aji]{whitehouse2023llm}
Chenxi Whitehouse, Monojit Choudhury, and Alham~Fikri Aji.
\newblock Llm-powered data augmentation for enhanced crosslingual performance.
\newblock \emph{arXiv preprint arXiv:2305.14288}, 2023.

\bibitem[Bai et~al.(2022)Bai, Jones, Ndousse, Askell, Chen, DasSarma, Drain,
  Fort, Ganguli, Henighan, et~al.]{bai2022training}
Yuntao Bai, Andy Jones, Kamal Ndousse, Amanda Askell, Anna Chen, Nova DasSarma,
  Dawn Drain, Stanislav Fort, Deep Ganguli, Tom Henighan, et~al.
\newblock Training a helpful and harmless assistant with reinforcement learning
  from human feedback.
\newblock \emph{arXiv preprint arXiv:2204.05862}, 2022.

\bibitem[K{\"o}pf et~al.(2023)K{\"o}pf, Kilcher, von R{\"u}tte, Anagnostidis,
  Tam, Stevens, Barhoum, Duc, Stanley, Nagyfi, et~al.]{kopf2023openassistant}
Andreas K{\"o}pf, Yannic Kilcher, Dimitri von R{\"u}tte, Sotiris Anagnostidis,
  Zhi-Rui Tam, Keith Stevens, Abdullah Barhoum, Nguyen~Minh Duc, Oliver
  Stanley, Rich{\'a}rd Nagyfi, et~al.
\newblock Openassistant conversations--democratizing large language model
  alignment.
\newblock \emph{arXiv preprint arXiv:2304.07327}, 2023.

\bibitem[Chiang et~al.(2024)Chiang, Zheng, Sheng, Angelopoulos, Li, Li, Zhang,
  Zhu, Jordan, Gonzalez, et~al.]{chiang2024chatbot}
Wei-Lin Chiang, Lianmin Zheng, Ying Sheng, Anastasios~Nikolas Angelopoulos,
  Tianle Li, Dacheng Li, Hao Zhang, Banghua Zhu, Michael Jordan, Joseph~E
  Gonzalez, et~al.
\newblock Chatbot arena: An open platform for evaluating llms by human
  preference.
\newblock \emph{arXiv preprint arXiv:2403.04132}, 2024.

\bibitem[Touvron et~al.(2023)Touvron, Martin, Stone, Albert, Almahairi, Babaei,
  Bashlykov, Batra, Bhargava, Bhosale, et~al.]{touvron2023llama2}
Hugo Touvron, Louis Martin, Kevin Stone, Peter Albert, Amjad Almahairi, Yasmine
  Babaei, Nikolay Bashlykov, Soumya Batra, Prajjwal Bhargava, Shruti Bhosale,
  et~al.
\newblock Llama 2: Open foundation and fine-tuned chat models.
\newblock \emph{arXiv preprint arXiv:2307.09288}, 2023.

\bibitem[{\"U}st{\"u}n et~al.(2024){\"U}st{\"u}n, Aryabumi, Yong, Ko, D'souza,
  Onilude, Bhandari, Singh, Ooi, Kayid, et~al.]{ustun2024aya}
Ahmet {\"U}st{\"u}n, Viraat Aryabumi, Zheng-Xin Yong, Wei-Yin Ko, Daniel
  D'souza, Gbemileke Onilude, Neel Bhandari, Shivalika Singh, Hui-Lee Ooi, Amr
  Kayid, et~al.
\newblock Aya model: An instruction finetuned open-access multilingual language
  model.
\newblock \emph{arXiv preprint arXiv:2402.07827}, 2024.

\bibitem[Gala et~al.(2024)Gala, Jayakumar, Husain, M, Khan, Kanojia,
  Puduppully, Khapra, Dabre, Murthy, and Kunchukuttan]{gala2024airavata}
Jay Gala, Thanmay Jayakumar, Jaavid~Aktar Husain, Aswanth~Kumar M, Mohammed
  Safi Ur~Rahman Khan, Diptesh Kanojia, Ratish Puduppully, Mitesh~M. Khapra,
  Raj Dabre, Rudra Murthy, and Anoop Kunchukuttan.
\newblock Airavata: Introducing hindi instruction-tuned llm.
\newblock \emph{arXiv preprint arXiv: 2401.15006}, 2024.

\bibitem[Son et~al.(2024)Son, Lee, Kim, Kim, Muennighoff, Choi, Park, Yoo, and
  Biderman]{son2024kmmlu}
Guijin Son, Hanwool Lee, Sungdong Kim, Seungone Kim, Niklas Muennighoff,
  Taekyoon Choi, Cheonbok Park, Kang~Min Yoo, and Stella Biderman.
\newblock Kmmlu: Measuring massive multitask language understanding in korean.
\newblock \emph{arXiv preprint arXiv:2402.11548}, 2024.

\bibitem[Sengupta et~al.(2023)Sengupta, Sahu, Jia, Katipomu, Li, Koto, Afzal,
  Kamboj, Pandit, Pal, et~al.]{sengupta2023jais}
Neha Sengupta, Sunil~Kumar Sahu, Bokang Jia, Satheesh Katipomu, Haonan Li,
  Fajri Koto, Osama~Mohammed Afzal, Samta Kamboj, Onkar Pandit, Rahul Pal,
  et~al.
\newblock Jais and jais-chat: Arabic-centric foundation and instruction-tuned
  open generative large language models.
\newblock \emph{arXiv preprint arXiv:2308.16149}, 2023.

\bibitem[Group et~al.(2024)Group, Levine, Huang, Wang, Batista, Szymanska,
  Ding, Chou, Pessiot, Effendi, et~al.]{group2024rakutenai}
Rakuten Group, Aaron Levine, Connie Huang, Chenguang Wang, Eduardo Batista, Ewa
  Szymanska, Hongyi Ding, Hou~Wei Chou, Jean-Fran{\c{c}}ois Pessiot, Johanes
  Effendi, et~al.
\newblock Rakutenai-7b: Extending large language models for japanese.
\newblock \emph{arXiv preprint arXiv:2403.15484}, 2024.

\end{thebibliography}

\vspace{10pt}
{\large \textbf{About Two Platforms}}

Two Platforms (TWO) is a tech startup that aims to redefine Human-AI Interaction, and is at the forefront of the next generation of AI that is visual and immersive. TWO is building consumer AI apps and services powered by its proprietary Gen-AI models. TWO is headquartered in Silicon Valley with offices in Seoul and Mumbai.

\end{document}